\documentclass{article} % For LaTeX2e
\usepackage{nips15submit_e,times}
\usepackage{hyperref}
\usepackage{url}
\usepackage{mysymbols}
\usepackage{subcaption}
\usepackage{amsmath, amsthm, amssymb}
\usepackage{graphicx}
\usepackage{cite}
\usepackage{tikz}
\usepackage{comment}
\usetikzlibrary{shapes.geometric, arrows}
\tikzstyle{startstop} = [rectangle, rounded corners, minimum width=3cm, minimum height=1cm,text centered, draw=black, fill=white!30]
\tikzstyle{io} = [trapezium, trapezium left angle=70, trapezium right angle=110, minimum width=3cm, minimum height=1cm, text centered, draw=black, fill=white!30]
\tikzstyle{process} = [rectangle, minimum width=3cm, minimum height=1cm, text centered, draw=black, fill=white!30]
\tikzstyle{decision} = [diamond, minimum width=3cm, minimum height=0.5cm, text centered, draw=black, fill=white!30]
\tikzstyle{arrow} = [thick,->,>=stealth]
\tikzstyle{line} = [draw, thick, -latex']

\title{A Surrogate-based Generic Classifier for \\Chinese TV Series Reviews}

\author{
Yufeng Ma\\
Department of Computer Science\\
Virginia Tech\\
Blacksburg, VA 24060 \\
\texttt{yufengma@vt.edu} \\
\And
Long Xia \\
Department of Business Information Technology\\
Virginia Tech\\
Blacksburg, VA 24060 \\
\texttt{longxia1@vt.edu} \\
\And
Wenqi Shen \\
Department of Business Information Technology\\
Virginia Tech\\
Blacksburg, VA 24060 \\
\texttt{shenw@vt.edu} \\
\And
Mi Zhou \\
School of Management\\
Xi'an Jiaotong Univeristy\\
ShaanXi Province, China 710049 \\
\texttt{zhoumi@xjtu.edu.cn} \\
\And
Weiguo Fan \\
Department of Accounting and Information Systems\\
Virginia Tech\\
Blacksburg, VA 24060 \\
\texttt{wfan@vt.edu} \\
}
\nipsfinalcopy % Uncomment for camera-ready version

\begin{document}

\maketitle

\begin{abstract}
With the emerging of various online video platforms like Youtube, Youku and LeTV, online TV series' reviews become more and more important both for viewers and producers. Customers rely heavily on these reviews before selecting TV series, while producers use them to improve the quality. As a result, automatically classifying reviews according to different requirements evolves as a popular research topic and is essential in our daily life. In this paper, we focused on reviews of hot TV series in China and successfully trained generic classifiers based on eight predefined categories. The experimental results showed promising performance and effectiveness of its generalization to different TV series.
\end{abstract}

\section{Introduction}
With Web 2.0's development, more and more commercial websites, such as Amazon, Youtube and Youku, encourage users to post product reviews on their platforms \cite{IDS1, IDS2}. These reviews are helpful for both readers and product manufacturers. For example, for TV or movie producers, online reviews indicates the aspects that viewers like and/or dislike. This information facilitates producers' production process. When producing future films TV series, they can tailer their shows to better accommodate consumers' tastes. For manufacturers, reviews may reveal customers' preference and feedback on product functions, which help manufacturers to improve their products in future development. On the other hand, consumers can evaluate the quality of product or TV series based on online reviews, which help them make final decisions of whether to buy or watch it. However, there are thousands of reviews emerging every day. Given the limited time and attention consumers have, it is impossible for them to allocate equal amount of attention to all the reviews. Moreover, some readers may be only interested in certain aspects of a product or TV series. It's been a waste of time to look at other irrelevant ones. As a result, automatic classification of reviews is essential for the review platforms to provide a better perception of the review contents to the users.

Most of the existing review studies focus on product reviews in English. While in this paper, we focus on reviews of hot Chinese movies or TV series, which owns some unique characteristics. First, Table~\ref{boxo} shows Chinese movies' development \cite{boxoffice} in recent years. The growth of box office and viewers is dramatically high in these years, which provides substantial reviewer basis for the movie/TV series review data. Moreover, the State Administration of Radio Film and Television also announced that the size of the movie market in China is at the 2nd place right after the North America market. In \cite{boxoffice}, it also has been predicted that the movie market in China may eventually become the largest movie market in the world within the next 5-10 years. Therefore, it is of great interest to researchers, practitioners and investors to understand the movie market in China.

Besides flourishing of movie/TV series, there are differences of aspect focuses between product and TV series reviews. When a reviewer writes a movie/TV series review, he or she not only care about the TV elements like actor/actress, visual effect, dialogues and music, but also related teams consisted of director, screenwriter, producer, etc. However, with product reviews, few reviewers care about the corresponding backstage teams. What they do care and will comment about are only product related issues like drawbacks of the product functions, or which aspect of the merchandise they like or dislike. Moreover, most of recent researchers' work has been focused on English texts due to its simpler grammatical structure and less vocabulary, as compared with Chinese. Therefore, Chinese movie reviews not only provide more content based information, but also raise more technical challenges. With bloom of Chinese movies,  automatic classification of Chinese movie reviews is really essential and meaningful.

\begin{table}[ht]
\caption{Chinese Movies Box Office Statistics}\label{boxo}
\begin{center}
\begin{tabular}{lcc}
\multicolumn{1}{c}{\bf Year}  &\multicolumn{1}{c}{\bf Box office (million)} & \multicolumn{1}{c}{\bf \# of viewers (million)}
\\ \hline \\
2009		& $6206$		& 	$204$\\
2010		& $10172$	& 	$286$\\
2011		& $13115$ 	& 	$370$\\
2012		& $17073$	& 	$467$\\
2013 	& $21769$	&	$613$\\
2014 	& $29600$	&	$830$\\
2015 	& $44000$	&	$1256$
\end{tabular}
\end{center}
\end{table}

In this paper, we proposed several strategies to make our classifiers generalizable to agnostic TV series. First, TV series roles' and actors/actresses' names are substituted by generic tags like role\_i and player\_j, where i and j defines their importance in this movie. On top of such kind of words, feature tokens are further manipulated by feature selection techniques like DRC or $\chi^2$, in order to make it more generic. We also experimented with different feature sizes with multiple classifiers in order to alleviate overfitting with high dimension features.

The remainder of this paper is organized as follows. Section 2 describes some related work. Section 3 states our problem and details our proposed procedure of approaching the problem. In Section 4, experimental results are provided and discussed. Finally, the conclusions are presented in Section 5.

\section{Related Work}
Since we are doing supervised learning task with text input, it is related with work of useful techniques like feature selections and supervised classifiers. Besides, there are only public movie review datasets in English right now, which is different from our language requirement. In the following of this section, we will first introduce some existing feature selection techniques and supervised classifiers we applied in our approach. Then we will present some relevant datasets that are normally used in movie review domain.

\subsection{Feature selection}
Feature selection, or variable selection is a very common strategy applied in machine learning domain, which tries to select a subset of relevant features from the whole set. There are mainly three purposes behind this. Smaller feature set or features with lower dimension can help researchers to understand or interpret the model they designed more easily. With fewer features, we can also improve the generalization of our model through preventing overfitting, and reduce the whole training time.

Document Relevance Correlation(DRC), proposed by W.~Fan et al 2005 \cite{DRC}, is a useful feature selection technique. The authors apply this approach to profile generation in digital library service and news-monitoring. They compared DRC with other well-known methods like Robertson's Selection Value \cite{RSV}, and machine learning based ones like information gain\cite{MI}. Promising experimental results were shown to demonstrate the effectiveness of DRC as a feature selection in text field.

Another popular feature selection method is called $\chi^2$ \cite{chi2}, which is a variant of $\chi^2$ test in statistics that tries to test the independence between two events. While in feature selection domain, the two events can be interpreted as the occurrence of feature variable and a particular class. Then we can rank the feature terms with respect to the $\chi^2$ value. It has been proved to be very useful in text domain, especially with bag of words feature model which only cares about the appearance of each term.

\subsection{Supervised Classifier}
What we need is to classify each review into several generic categories that might be attractive to the readers, so classifier selection is also quite important in our problem. Supervised learning takes labeled training pairs and tries to learn an inferred function, which can be used to predict new samples. In this paper, our selection is based on two kinds of learning, i.e., discriminative and generative learning algorithms. And we choose three typical algorithms to compare. \Naive Bayes \cite{NB}, which is the representative of generative learning, will output the class with the highest probability that is generated through the bayes' rule. While for the discriminative classifiers like logistic regression \cite{LR} or Support Vector Machine \cite{SVM}, final decisions are based on the classifier's output score, which is compared with some threshold to distinguish between different classes.

\subsection{TV series Review Dataset}
Dataset is another important factor influencing the performance of our classifiers. Most of the public available movie review data is in English, like the IMDB dataset collected by Pang/Lee 2004 \cite{IMDB}. Although it covers all kinds of movies in IMDB website, it only has labels related with the sentiment. Its initial goal was for sentiment analysis. Another intact movie review dataset is SNAP \cite{SNAP}, which consists of reviews from Amazon but only bearing rating scores. However, what we need is the content or aspect tags that are being discussed in each review. In addition, our review text is in Chinese. Therefore, it is necessary for us to build the review dataset by ourselves and label them into generic categories, which is one of as one of the contributions of this paper.

\section{Chinese TV series Review Classification}
\begin{figure*}[htb]
\begin{center}
\begin{tikzpicture}[node distance=2cm]

\node (in1) [io]{{\color{white}{xx}}Douban{\color{white}{xx}}};
\node (in2) [io, right of=in1, xshift=4cm]{Baidu Encyclopedia};

\node (pro1) [process, below of=in1]{TV series Reviews};
\node (pro)   [process, below of=in2]{Role/actor Knowledgebase};

\node (pro2) [process, below of=pro]{Generic Movie Review};
\node (pro3) [process, below of=pro2]{Text Tokens wth Surrogate};
\node (pro4) [process, below of=pro3]{Reduced Features};
\node (pro5) [process, below of=pro4]{NB/LR/SVM};
\node (pro6) [process, below of=pro1, yshift=-100]{Topics};
\node (out7) [io, below of=pro5]{Generic Classifiers};
\node (pro7) [process, below of=pro1, yshift=-40]{Raw Text Tokens};

\draw [arrow] (in2) -- node[anchor=west]{knowledge crawler} (pro);
\draw [arrow] (in1) -- node[anchor=east]{review crawler} (pro1);
\draw [arrow] (pro) -- node[anchor=east]{Substitution} (pro2);
\draw [arrow] (pro1) |- (pro2);
\draw [arrow] (pro2) -- node[anchor=west]{Tokenization/Stop words removal} (pro3);
\draw [arrow] (pro3) -- node[anchor=east]{DRC and $\chi^2$} (pro4);
\draw [arrow] (pro4) -- (pro5);
\draw [arrow] (pro1) -- (pro7);
\draw [arrow] (pro6) |- node[anchor=north]{Category labels}(pro5);
\draw [arrow] (pro5) -- (out7);
\draw [arrow] (pro7) -- node[anchor=east]{LDA}(pro6);

\end{tikzpicture}
\end{center}

\caption{Procedure of Building Generic Classifiers}\label{procedure}
\end{figure*}
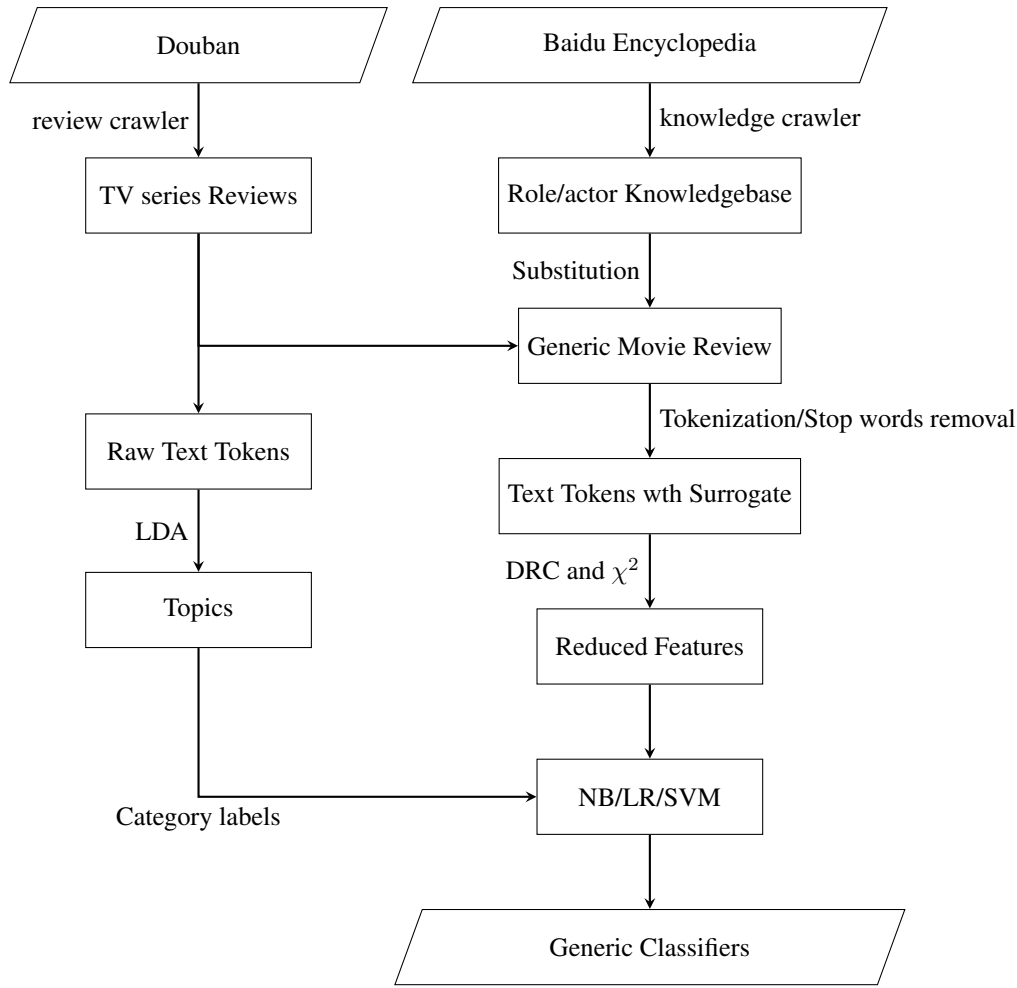

Let $R=r_1, r_2, \ldots, r_n$ be a set of Chinese movie reviews with no categorical information. The ultimate task of movie review classification is to label them into different predefined categories as $c_1, c_2, \ldots, c_m$. Starting from scratch, we need to collect such review set $R$ from an online review website and then manually label them into generic categories $\{c_i\}$. Based on the collected dataset, we can apply natural language processing techniques to get raw text features and further learn the classifiers. In the following subsections, we will go through and elaborate all the subtasks shown in Figure~\ref{procedure}.

\subsection{Building Dataset}
What we are interested in are the reviews of the hottest or currently broadcasted TV series, so we select one of the most influential movie and TV series sharing websites in China, Douban. For every movie or TV series, you can find a corresponding section in it. For the sake of popularity, we choose ``The Journey of Flower'', ``Nirvana in Fire'' and ``Good Time'' as parts of our movie review dataset, which are the hottest TV series from summer to fall 2015. Reviews of each episode have been collected for the sake of dataset comprehensiveness.

Then we built the crawler written in python with the help of scrapy. Scrapy will create multiple threads to crawl information we need simultaneously, which saves us lots of time. For each episode, it collected both the short description of this episode and all the reviews under this post. The statistics of our TV series review dataset is shown in Table~\ref{statData}.
\begin{table}[ht]
\caption{Statistics of TV series Review Dataset}\label{statData}
\begin{center}
\begin{tabular}{ll}
\multicolumn{1}{c}{\bf TV series}  &\multicolumn{1}{c}{\bf \# of reviews}
\\ \hline \\
The Journey of Flower             & $6381$\\
Nirvana in Fire             &$7527$ \\
Good Time             & $5424$
\end{tabular}
\end{center}
\end{table}

\subsection{Basic Text Processing}
Based on the collected reviews, we are ready to build a rough classifier. Before feeding the reviews into a classifier, we applied two common procedures: tokenization and stop words removal for all the reviews. We also applied a common text processing technique to make our reviews more generic. We replaced the roles' and actors/actresses' names in the reviews with some common tokens like role\_i, actor\_j, where i and j are determined by their importance in this TV series. Therefore, we have the following inference
\begin{align}
i < j \Longrightarrow IM(i) > IM(j)
\end{align}
where $IM(*)$ is a function which map a role's or actor's index into its importance. However, in practice, it is not a trivial task to infer the importance of all actors and actresses. We rely on data onBaidu Encyclopedia, which is the Chinese version of Wikipedia. For each movie or TV series, Baidu Encyclopedia has all the required information, which includes the level of importance for each role and actor in the show. Actor/actress in a leading role will be listed at first, followed by the ones in a supporting role and other players. Thus we can build a crawler to collect such information, and replace the corresponding words in reviews with generic tags. 

Afterwards, word sequence of each review can be manipulated with tokenization and stop words removal. Each sequence is broken up into a vector of unigram-based tokens using NLPIR \cite{NLPIR}, which is a very powerful tool supporting sentence segmentation in Chinese. Stop words are words that do not contribute to the meaning of the whole sentence and are usually filtered out before following data processing. Since our reviews are collected from online websites which may include lots of forum words, for this particular domain, we include common forum words in addition to the basic Chinese stop words. Shown below are some typical examples in English that are widely used in Chinese forums.
\begin{align*}
\mbox{BBS, BT, NB, BS, CU, LOL, 4242, SF, YY, \ldots}
\end{align*}
These two processes will help us remove significant amount of noise in the data.

\subsection{Topic Modelling and Labeling}
With volumes of TV series review data, it's hard for us to define generic categories without looking at them one by one. Therefore, it's necessary to run some unsupervised models to get an overview of what's being talked in the whole corpus. Here we applied Latent Dirichlet Allocation \cite{LDA, LDAdesign}  to discover the main topics related to the movies and actors. In a nutshell, the LDA model assumes that there exists a hidden structure consisting of the topics appearing in the whole text corpus. The LDA algorithm uses the co-occurrence of observed words to learn this hidden structure. Mathematically, the model calculates the posterior distribution of the unobserved variables. Given a set of training documents, LDA will return two main outputs. The first is the list of topics represented as a set of words, which presumably contribute to this topic in the form of their weights. The second output is a list of documents with a vector of weight values showing the probability of a document containing a specific topic.

Based on the results from LDA, we carefully defined eight generic categories of movie reviews which are most representative in the dataset as shown in Table~\ref{CMR}.

\begin{table}[htb]
\centering
\caption{Categories of Movie Reviews}\label{CMR}
\begin{tabular}{|c|c|}
	\hline
Categories &   Specific Meaning   \\
	\hline
Plot of the TV series& development of the plot in TV series\\
Actor/actress & actors related with this TV series\\ 
Role & like or dislike specific roles, or role related\\ 
Dialogue &  discussion or analysis of the dialogues\\
Analysis & deep analysis of the plot and role's inner part activity\\  
Platform & related with specific platform like audience rate on some Channels\\
Thumb up or down & simply follow the post with common forum words\\
Noise or others & advertisements or ones making nonsense\\
	\hline
\end{tabular}
\end{table}
The purpose of this research is to classify each review into one of the above $8$ categories. In order to build reasonable classifiers, first we need to obtain a labeled dataset. Each of the TV series reviews was labeled by at least two individuals, and only those reviews with the same assigned label were selected in our training and testing data. This approach ensures that reviews with human biases are filtered out. As a result, we have $5000$ for each TV series that matches the selection criteria.

\subsection{Feature Selection}
After the labelled cleaned data has been generated, we are now ready to process the dataset. One problem is that the vocabulary size of our corpus will be quite large. This could result in overfitting with the training data. As the dimension of the feature goes up, the complexity of our model will also increase. Then there will be quite an amount of difference between what we expect to learn and what we will learn from a particular dataset. One common way of dealing with the issue is to do feature selection. Here we applied DRC and $\chi^2$ mentioned in related work. First let's define a contingency table for each word $j$ like in Table~\ref{cont}. If $j=1$, it means the appearance of word~$j$.
\begin{table}[ht]
\caption{Contingency Table for Word $j$}\label{cont}
\begin{center}
\begin{tabular}{cccc}
\hline\\
\multicolumn{1}{c}{} & \multicolumn{1}{c}{\bf Relevant}  &\multicolumn{1}{c}{\bf Irrelevant} & \multicolumn{1}{c}{}
\\ \hline \\
Word $j=1$ & A             &B & A+B\\
Word $j=0$ & C             & D & C+D\\
& A+C & B+D & N\\
\hline
\end{tabular}
\end{center}
\end{table}

Recall that in classical statistics, $\chi^2$ is a method designed to measure the independence between two variables or events, which in our case is the word $j$ and its relevance to the class $i$. Higher $\chi^2$ value means higher correlations between them. Therefore, based on the definition of $\chi^2$ in \cite{chi2} and the above Table~\ref{cont}, we can represent the $\chi^2$ value as below:
\begin{align}
\chi^2 = \frac{N \times (AD - CB)}{(A+C) \times (B+D) \times (A+B) \times (C+D) }
\end{align}
While for DRC method, it's based on Relevance Correlation Value, whose purpose is to measure the similarity between two distributions, i.e., binary distribution of word $j$'s occurrence and documents' relevance to class $i$ along all the training data. For a particular word $j$, its occurrence distribution along all the data can be represented as below (assume we have $N$ reviews):
\begin{align}
I_j = \langle 1, 1, 0, \ldots, 0, 1 \rangle
\end{align}
And we also know each review $r_k$'s relevance with respect to $c_i$ using the manually tagged labels.
\begin{align}
Rel_k = \langle 1, 0, 0, \ldots, 1, 0 \rangle
\end{align}
where $0$ means irrelevant and $1$ means relevant. Therefore, we can calculate the similarity between these two vectors as 
\begin{align}
\mbox{RCV}_j = \frac{\sum_{k=1}^N I_{kj} Rel_k}{ \sqrt{\sum_{k=1}^N I_{kj}^2 } \sqrt{ \sum_{k=1}^N Rel_k^2 } }
\end{align}
where $\mbox{RCV}_j$ is called the Relevance Correlation Value for word $j$. Because $I_{ij}$ is either $1$ or $0$, with the notation in the contingency table, RCV can be simplified as 
\begin{align}
\mbox{RCV}_j = \frac{A}{\sqrt{A+B} \sqrt{A+C}}
\end{align}
Then on top of RCV, they incorporate the probability of the presence of word $j$ if we are given that the document is relevant. In this way, our final formula for computing DRC becomes
\begin{align}
\mbox{DRC}_j &= \quad P(w_j=1|R) \cdot RCV_j \\
&= \quad \frac{A^2}{\sqrt{A+B}}.
\end{align}

Therefore, we can apply the above two methods to all the word terms in our dataset and choose words with higher $\chi^2$ or DRC values to reduce the dimension of our input features.

\subsection{Learning Classifiers}
Finally, we are going to train classifiers on top of our reduced generic features. As mentioned above, there are two kinds of learning algorithms, i.e., discriminant and generative classifiers. Based on Bayes rule, the optimal classifier is represented as 
\begin{align*}
h^*( \textbf{x} )\quad &=\quad \arg\max_{c_i} \log p(y=c_i|\textbf{x}) \\
& \propto\quad \arg\max_{c_i} \{ \log p(\textbf{x} | y=c_i) + \log p(y=c_i)\}
\end{align*}
where $p(y=c_i)$ is the prior information we know about class $c_i$. 

So for generative approach like \Naive Bayes, it will try to estimate both $p(\textbf{x}|y)$ and $p(y)$. During testing time, we can just apply the above Bayes rule to predict $y$. Why do we call it naive? Remember that we assume that each feature is conditionally independent with each other. So we have
\begin{align}
p(\textbf{x}|y) = p(x_1|y) \cdot p(x_2 | y) \ldots p(x_{m-1}|y) \cdot p(x_m | y)
\end{align}
where we made the assumption that there are $m$ words being used in our input. If features are binary, for each word $j$ we may simply estimate the probability by 
\begin{align}
p(x_i=1|y=c_i) = \frac{ \#\{ x_i=1, y=c_i\} + l }{ \#\{y=c_i\} + 2l}
\end{align}
in which, $l$ is a smoothing parameter in case there is no training sample for $y=c_i$ and $\#\{\ast\}$ outputs the number of a set. With all these probabilities computed, we can make decisions by whether
\begin{align}
\log \frac{p(y=c_i|\textbf{x})}{p(y\ne c_i | \textbf{x})} = \sum_{k=1}^m \log\frac{p(x_k|y=c_i)}{p(x_k | y\ne c_k)} + \log\frac{p(y=c_i)}{p(y\ne c_i)}\ge 0
\end{align}

On the other hand, discriminant learning algorithms will estimate $p(y|\textbf{x})$ directly, or learn some ``discriminant'' function $h(\textbf{x})$. Then by comparing $h(\textbf{x})$ with some threshold, we can make the final decision. Here we applied two common classifiers logistic regression and support vector machine to classify movie reviews. Logistic regression squeezes the input feature into some interval between $0$ and $1$ by the sigmoid function, which can be treated as the probability $p(y|\textbf{x})$. 
\begin{align}\label{LRE}
p(y=c_i|\textbf{x}) &= \sigma( \textbf{w}^T \textbf{x} ) = \frac{1}{1 + e^{-w_0 - \sum_i w_i x_i}}\\
p(y\ne c_i|\textbf{x}) &= 1- \sigma( \textbf{w}^T \textbf{x} ) = \frac{e^{-w_0 - \sum_i w_i x_i}}{1 + e^{-w_0 - \sum_i w_i x_i}}
\end{align}
The Maximum A Posteriori of logistic regression with Gaussian priors on parameter $\textbf{w}$ is defined as below
\begin{align*}
\hat{\textbf{w}} &= \arg\max_{\textbf{w}} \mbox{Loss}(\textbf{w})= \arg\max_{\textbf{w}} \log P(\textbf{w}|\{\textbf{x}_k, y_k\}) \\
&\propto \arg\max_{\textbf{w}} \log P(Y| X, \textbf{w}) P(\textbf{w}) \\
& = \arg\max_{\textbf{w}} \sum_{k=1}^N \log P(Y_k = y_k | \{\textbf{x}_k, \textbf{w} ) + \sum_{j=0}^m\log P(w_j)
\end{align*}
which is a concave function with respect to $\textbf{w}$, so we can use gradient ascent below to optimize the objective function and get the optimal $\textbf{w}$.
\begin{align}
\textbf{w}^{(t+1)} = \textbf{w}^{(t)}  + \eta \frac{\partial Loss(\textbf{w})}{\partial \textbf{w}}
\end{align}
where $\eta$ is a positive hyper parameter called learning rate. Then we can just use equation~(\ref{LRE}) to distinguish between classes.

While for Support Vector Machine(SVM), its initial goal is to learn a hyperplane, which will maximize the margin between the two classes' boundary hyperplanes. Suppose the hyperplane we want to learn is 
\begin{align*}
\textbf{w}^T\textbf{x} + w_0 = 0
\end{align*}
Then the soft-margin version of SVM is
\begin{align*}
\min_{\textbf{w}, w_0, \xi_k} &\quad \frac{1}{2} \textbf{w}^T \textbf{w} + C \sum_{k=1}^N \xi_k \\
\st&\quad y_k\cdot(\textbf{w}^T\textbf{x}_k + w_0 ) \ge 1 - \xi_k\\
& \quad \xi_k \ge 0 
\end{align*}
where $\xi_k$ is the slack variable representing the error w.r.t. datapoint $(x_k, y_k)$. If we represent the inequality constraints by hinge loss function
\begin{align}
h(z) = \max\{0, 1 - z\},
\end{align}
What we want to minimize becomes
\begin{align}
\min_{\textbf{w}, w_0, \xi_k} &\quad \frac{1}{2} \textbf{w}^T \textbf{w} + C \sum_{k=1}^N h[ y_k\cdot(\textbf{w}^T\textbf{x}_k + w_0  ) ]
\end{align}
which can be solved easily with a Quadratic Programming solver. With learned $\textbf{w}$ and $w_0$, decision is made by determining whether 
\begin{align}
\textbf{w}^T x + w_0 \ge 0.
\end{align}

Based on these classifiers, we may also apply some kernel trick function on input feature to make originally linearly non-separable data to be separable on mapped space, which can further improve our classifier performance. What we've tried in our experiments are the polynomial and rbf kernels.

\section{Experimental Results and Discussion}
As our final goal is to learn a generic classifier, which is agnostic to TV series but can predict review's category reasonably, we did experiments following our procedures of building the classifier as discussed in section 1.

\subsection{Category Determining by LDA}
\begin{figure}[ht]
\centering
\includegraphics[width=0.9\textwidth]{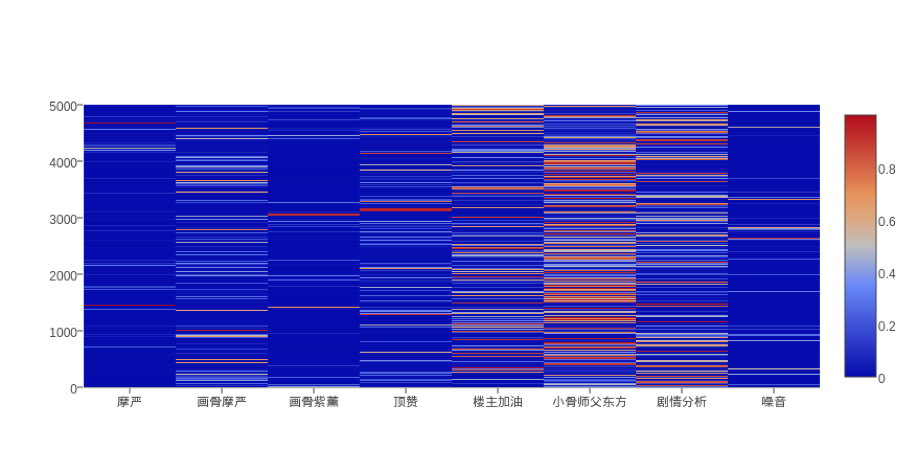}
\caption{LDA results with 8 topics}
\label{lda8}
\end{figure}

Before defining the categories of the movie reviews, we should first run some topic modeling method. Here we define categories with the help of LDA. With the number of topics being set as eight, we applied LDA on ``The Journey of Flower'', which is the hottest TV series in 2015 summer. As we rely on LDA to guide our category definition, we didn't run it on other TV series. The results are shown in Figure~\ref{lda8}. Note that the input data here haven't been replaced with the generic tag like role\_i or actor\_j, as we want to know the specifics being talked by reviewers. Here we present it in the form of heat maps. For lines with brighter color, the corresponding topic is discussed more, compared with others on the same height for each review. As the original texts are in Chinese, the output of LDA are represented in Chinese as well. 

We can see that most of the reviews are focused on discussing the roles and analyzing the plots in the movie, i.e., 6th and 7th topics in Figure~\ref{lda8}, while quite a few are just following the posts, like the 4th and 5th topic in the figure. Based on the findings, we generate the category definition shown in Table~\ref{CMR}. Then $5000$ out of each TV series reviews, with no label bias between readers, are selected to make up our final data set.

\subsection{Feature Size Comparison}
\begin{figure*}
\centering

\begin{subfigure}[t]{0.45\textwidth}
\centering
\includegraphics[width=\textwidth]{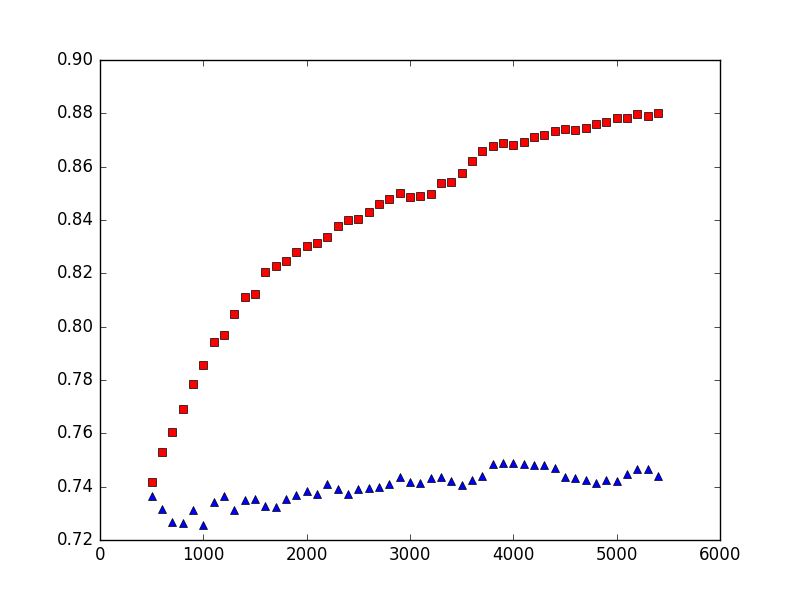}
\caption{Plot}
\end{subfigure}%
\hfill
\begin{subfigure}[t]{0.45\textwidth}
\centering
\includegraphics[width=\textwidth]{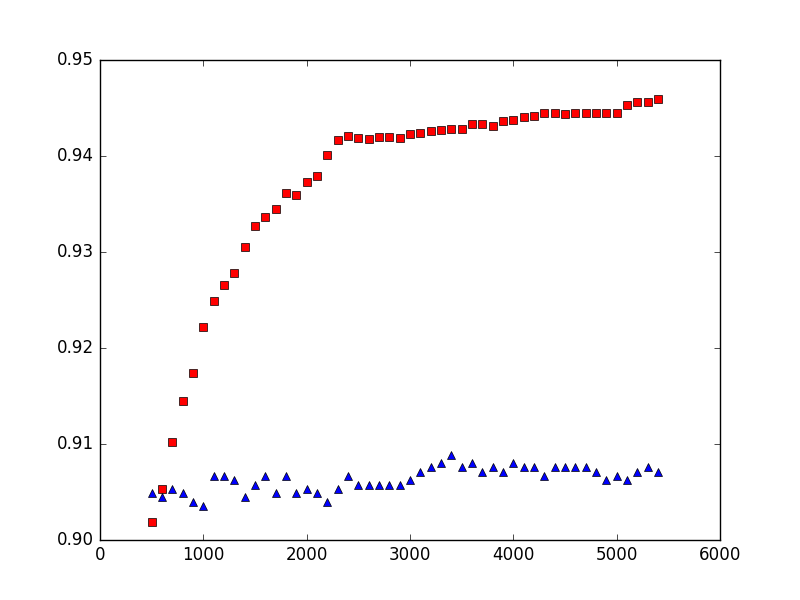}
\caption{Actor/actress}
\end{subfigure}

\bigskip 

\begin{subfigure}[t]{0.45\textwidth}
\centering
\includegraphics[width=\textwidth]{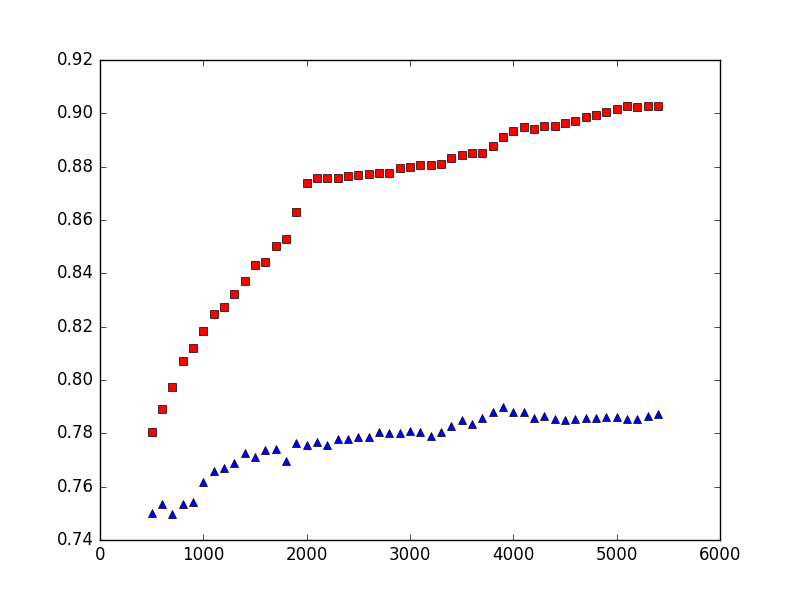}
\caption{Role}
\end{subfigure}%
\hfill
\begin{subfigure}[t]{0.45\textwidth}
\centering
\includegraphics[width=\textwidth]{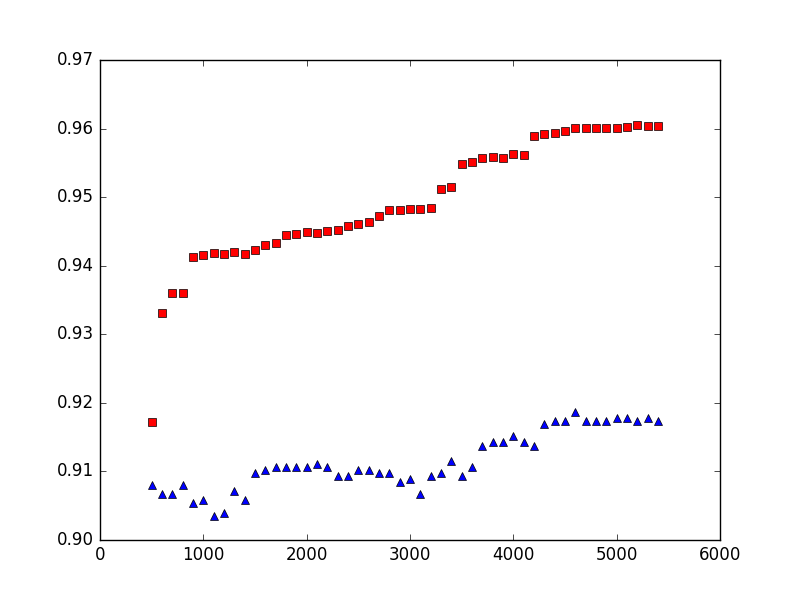}
\caption{Dialogue}
\end{subfigure}

\begin{subfigure}[t]{0.45\textwidth}
\centering
\includegraphics[width=\textwidth]{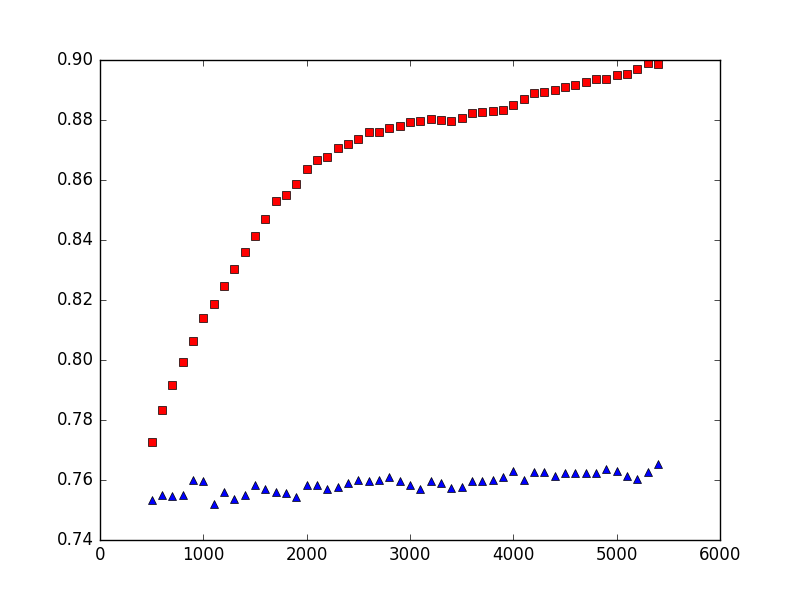}
\caption{Analysis}
\end{subfigure}%
\hfill
\begin{subfigure}[t]{0.45\textwidth}
\centering
\includegraphics[width=\textwidth]{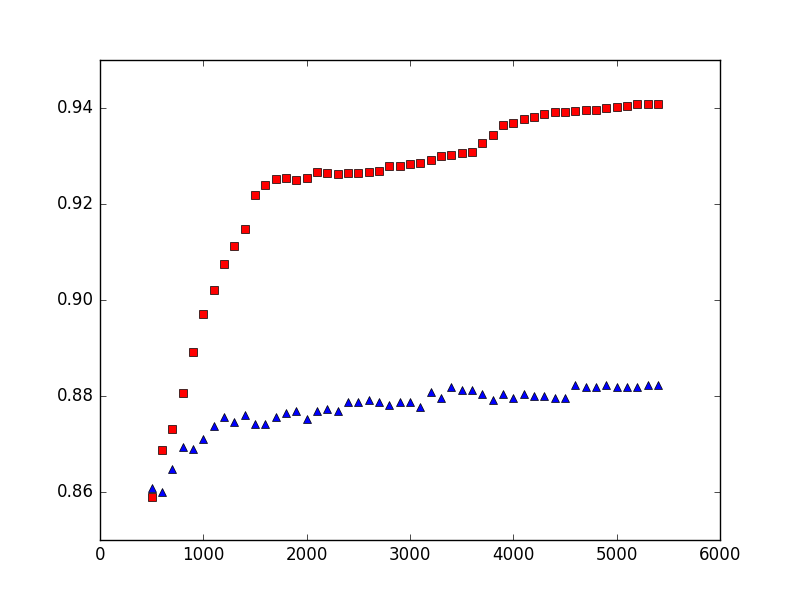}
\caption{Platform}
\end{subfigure}

\begin{subfigure}[t]{0.45\textwidth}
\centering
\includegraphics[width=\textwidth]{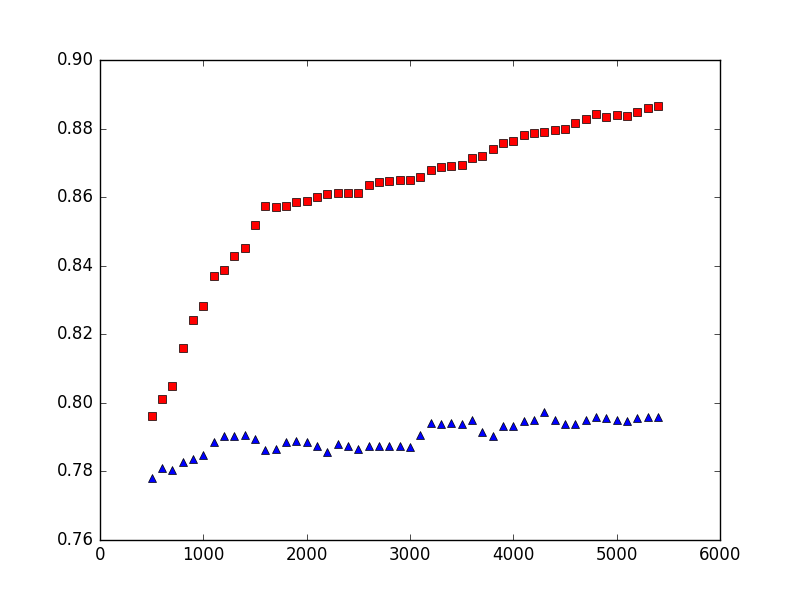}
\caption{Thumb up or down}
\end{subfigure}%
\hfill
\begin{subfigure}[t]{0.45\textwidth}
\centering
\includegraphics[width=\textwidth]{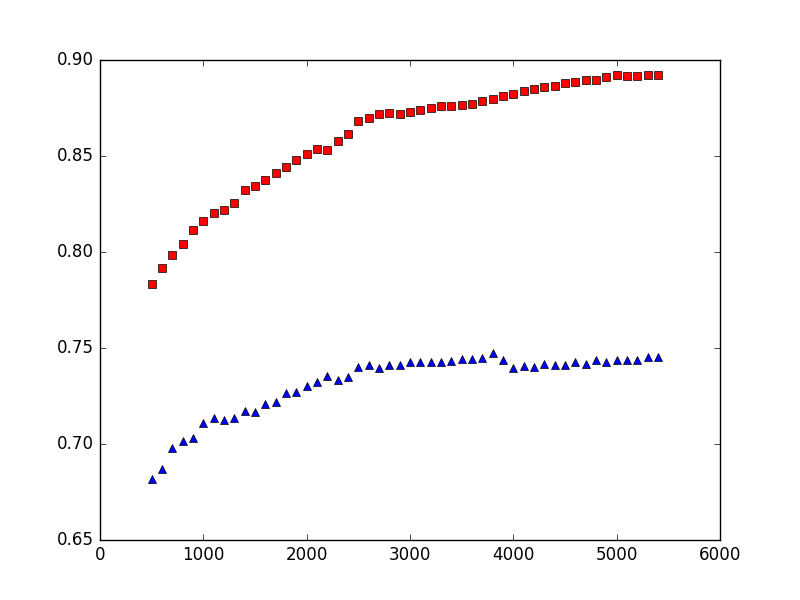}
\caption{Noise}
\end{subfigure}

\caption{Accuracy vs Feature size on 8 classifiers}
\label{fs}
\end{figure*}

Based on $\chi^2$ and DRC discussed in section~3.4, we can sort the importance of each word term. With different feature size, we can train the eight generic classifiers and get their performances on both training and testing set. Here we use SVM as the classifier to compare feature size's influence. Our results suggest that it performs best among the three. The results are shown in Figure~\ref{fs}. The red squares represent the training accuracy, while the blue triangles are testing accuracies.

As shown in Figure~\ref{fs}, it is easy for us to determine the feature size for each classifier. Also it's obvious that test accuracies of classifiers for plot, actor/actress, analysis, and thumb up or down, didn't increase much with adding more words. Therefore, the top $1000$ words with respect to these classes are fixed as the final feature words. While for the rest of classifiers, they achieved top testing performances at the size of about $4000$. Based on these findings, we use different feature sizes in our final classifiers.

\subsection{Generalization of Classifiers}
To prove the generalization of our classifiers, we use two of the TV series as training data and the rest as testing set. We compare them with classifiers trained without the replacement of generic tags like role\_i or actor\_j. So 3 sets of experiments are performed, and each are trained on top of \Naive Bayes, Logistic Regression and SVM. Average accuracies among them are reported as the performance measure for the sake of space limit. The results are shown in Table~\ref{GCT}. ``1'', ``2'' and ``3'' represent the TV series ``The Journey of Flower'', ``Nirvana in Fire'' and ``Good Time'' respectively. In each cell, the left value represents accuracy of classifier without replacement of generic tags and winners are bolded.

\begin{table}[htb]
\centering
\caption{Performance of 8 Classifiers}\label{GCT}
\begin{tabular}{|l|c|c|c|}
	\hline
Classifiers &   1\&2 - 3 & 1\&3 - 2 & 2 \& 3 - 1   \\
	\hline
Plot of the TV series & 72.33\%, \textbf{74.17\%} & 70.23\%, \textbf{72.78\%} & 74.22\%, \textbf{75.12\%} \\
Actor/actress &  87.23\%, \textbf{89.03\%} &  88.38\%, \textbf{89.44\%} &  90.23\%, \textbf{91.33\%} \\ 
Role & 85.33\%, \textbf{86.57\%} & 84.29\%, \textbf{86.37\%} & 85.22\%, \textbf{87.55\%} \\ 
Dialogue & 93.27\%, \textbf{94.11\%} & 92.56\%, \textbf{93.01\%} & 94.23\%, \textbf{94.85\%} \\
Analysis & 80.22\%,\textbf{ 81.38\%} & 79.29\%, \textbf{80.45\%} & 81.38\%, \textbf{82.11\%} \\  
Platform & \textbf{90.38\%}, 90.22\% & \textbf{89.47\%}, 89.45\% & \textbf{91.22\%}, 90.48\% \\
Thumb up or down & 83.77\%, 83.77\% & 84.25\%, \textbf{84.37\%} & 82.11\%, \textbf{83.21\%} \\
Noise or others & \textbf{86.11\%}, 85.28\% & \textbf{85.34\%}, 84.16\% & \textbf{84.58\%}, 83.79\% \\
	\hline
\end{tabular}
\end{table}
From the above table, we can see with substitutions of generic tags in movie reviews, the top 5 classifiers have seen performance increase, which indicates the effectiveness of our method. However for the rest three classifiers, we didn't see an improvement and in some cases the performance seems decreased. This might be due to the fact that in the first five categories, roles' or actors' names are mentioned pretty frequently while the rest classes don't care much about these. But some specific names might be helpful in these categories' classification, so the performance has decreased in some degree.

\section{Conclusion}
In this paper, a surrogate-based approach is proposed to make TV series review classification more generic among reviews from different TV series. Based on the topic modeling results, we define eight generic categories and manually label the collected TV series' reviews. Then with the help of Baidu Encyclopedia, TV series' specific information like roles' and actors' names are substituted by common tags within TV series domain. Our experimental results showed that such strategy combined with feature selection did improve the performance of classifications. Through this way, one may build classifiers on already collected TV series reviews, and then successfully classify those from new TV series. Our approach has broad implications on processing movie reviews as well. Since movie reviews and TV series reviews share many common characteristics, this approach can be easily applied to understand movie reviews and help movie producers to better process and classify consumers' movie review with higher accuracy.

%\section{Acknowledgement}

\bibliography{MovieReview} 

\begin{thebibliography}{10}

\bibitem{IDS1}
K.~Munson, H.~H. Thompson, J.~Cabaniss, H.~Nance, P.~Erlandsen, M.~McGrath, and
  M.~McGrath, ``The world is your library, or the state of international
  interlibrary loan in 2015,'' {\em Interlending \& Document Supply}, vol.~44,
  no.~2, 2016.

\bibitem{IDS2}
M.~Goldner and K.~Birch, ``Resource sharing in a cloud computing age,'' {\em
  Interlending \& Document Supply}, vol.~40, no.~1, pp.~4--11, 2012.

\bibitem{boxoffice}
C.~Film, ``Chinese movie big data analysis report.''
  \url{http://sanwen8.cn/p/197jW1H.html}, 2016 (accessed November 5, 2016).

\bibitem{DRC}
W.~Fan, M.~D. Gordon, and P.~Pathak, ``Effective profiling of consumer
  information retrieval needs: a unified framework and empirical comparison,''
  {\em Decision Support Systems}, vol.~40, no.~2, pp.~213--233, 2005.

\bibitem{RSV}
S.~E. Robertson, ``On relevance weight estimation and query expansion,'' {\em
  Journal of Documentation}, vol.~42, no.~3, pp.~182--188, 1986.

\bibitem{MI}
A.~Kraskov, H.~St{\"o}gbauer, R.~G. Andrzejak, and P.~Grassberger,
  ``Hierarchical clustering based on mutual information,'' {\em arXiv preprint
  q-bio/0311039}, 2003.

\bibitem{chi2}
F.~Yates, ``Contingency tables involving small numbers and the $\chi$ 2 test,''
  {\em Supplement to the Journal of the Royal Statistical Society}, vol.~1,
  no.~2, pp.~217--235, 1934.

\bibitem{NB}
J.~D. Rennie, L.~Shih, J.~Teevan, D.~R. Karger, {\em et~al.}, ``Tackling the
  poor assumptions of naive bayes text classifiers,'' in {\em ICML}, vol.~3,
  pp.~616--623, Washington DC, 2003.

\bibitem{LR}
S.~H. Walker and D.~B. Duncan, ``Estimation of the probability of an event as a
  function of several independent variables,'' {\em Biometrika}, vol.~54,
  no.~1-2, pp.~167--179, 1967.

\bibitem{SVM}
C.~Cortes and V.~Vapnik, ``Support-vector networks,'' {\em Machine learning},
  vol.~20, no.~3, pp.~273--297, 1995.

\bibitem{IMDB}
B.~Pang and L.~Lee, ``A sentimental education: Sentiment analysis using
  subjectivity summarization based on minimum cuts,'' in {\em Proceedings of
  the 42nd annual meeting on Association for Computational Linguistics},
  p.~271, Association for Computational Linguistics, 2004.

\bibitem{SNAP}
J.~J. McAuley and J.~Leskovec, ``From amateurs to connoisseurs: modeling the
  evolution of user expertise through online reviews,'' in {\em Proceedings of
  the 22nd international conference on World Wide Web}, pp.~897--908, ACM,
  2013.

\bibitem{NLPIR}
L.~Zhou and D.~Zhang, ``Nlpir: A theoretical framework for applying natural
  language processing to information retrieval,'' {\em Journal of the American
  Society for Information Science and Technology}, vol.~54, no.~2,
  pp.~115--123, 2003.

\bibitem{LDA}
D.~M. Blei, A.~Y. Ng, and M.~I. Jordan, ``Latent dirichlet allocation,'' {\em
  Journal of machine Learning research}, vol.~3, no.~Jan, pp.~993--1022, 2003.

\bibitem{LDAdesign}
S.~Moghaddam and M.~Ester, ``On the design of lda models for aspect-based
  opinion mining,'' in {\em Proceedings of the 21st ACM international
  conference on Information and knowledge management}, pp.~803--812, ACM, 2012.

\end{thebibliography}
\bibliographystyle{ieeetr}

\end{document}